\def\year{2022}\relax
\documentclass[letterpaper]{article} 
\usepackage{aaai22}  
\usepackage{times}  
\usepackage{helvet}  
\usepackage{courier}  
\usepackage[hyphens]{url}  
\usepackage{graphicx} 
\usepackage{natbib}  
\usepackage{caption} 
\DeclareCaptionStyle{ruled}{labelfont=normalfont,labelsep=colon,strut=off} 
\usepackage{latexsym}
\usepackage{amsmath,amssymb,amsfonts}
\usepackage{textcomp}
\usepackage{url}
\usepackage{comment}
\usepackage{subcaption}
\usepackage{multirow}
\usepackage{tabularx}
\usepackage{txfonts}
\usepackage{pifont}
\usepackage{algcompatible}
\usepackage{relsize}
\usepackage{stmaryrd}
\usepackage{epsfig}
\usepackage{graphicx}
\usepackage{pifont}

\usepackage[ruled]{algorithm2e}
\usepackage{algorithmicx}
\usepackage{algpseudocode}
\usepackage{bbm}

\usepackage{tikz}
\usepackage{listings}
\usetikzlibrary{decorations.pathreplacing,calc}
\newcommand{\tikzmark}[1]{\tikz[overlay,remember picture] \node (#1) {};}

\newcommand*{\AddNote}[4]{%
    \begin{tikzpicture}[overlay, remember picture]
        \draw [decoration={brace,amplitude=0.5em},decorate,thick,black]
            ($(#3)!([yshift=1.5ex]#1)!($(#3)-(0,1)$)$) --  
            ($(#3)!(#2)!($(#3)-(0,1)$)$)
                node [align=center, text width=2.5cm, pos=0.5, anchor=west] {#4};
    \end{tikzpicture}
}%

\newcommand*{\AddNotee}[4]{%
    \begin{tikzpicture}[overlay, remember picture]
        \draw [decoration={brace,amplitude=0.2em},decorate, thick,black]
            ($(#3)!([yshift=1.5ex]#1)!($(#3)-(0,1)$)$) --  
            ($(#3)!(#2)!($(#3)-(0,1)$)$)
                node [align=center, text width=2.5cm, pos=0.5, anchor=west] {#4};
    \end{tikzpicture}
}%

\title{Dual Task Framework for Improving Persona-grounded Dialogue Dataset}
\author {
    Minju Kim\thanks{The authors contribute equally to this paper.}\textsuperscript{\rm 1},
    Beong-woo Kwak$^*$\textsuperscript{\rm 1}, 
    Youngwook Kim\textsuperscript{\rm 1}, 
    Hong-in Lee\textsuperscript{\rm 1} \\
    Seung-won Hwang\textsuperscript{\rm 2} and 
    Jinyoung Yeo\thanks{Corresponding author (Email: jinyeo@yonsei.ac.kr)}\textsuperscript{\rm 1}
}
\affiliations {
    \textsuperscript{\rm 1}Yonsei University \textsuperscript{\rm 2}Seoul National University
}

\usepackage{bibentry}

\begin{document}

\maketitle


\newtheorem{example}{Example}

\newcommand\Tstrut{\rule{0pt}{2.2ex}}       
\newcommand\Bstrut{\rule[-0.6ex]{0pt}{0pt}} 
\newcommand{\TBstrut}{\Tstrut\Bstrut} 

\newcommand{\se}{{\it SE}}%
\newcommand{\eg}{{\it e.g.}}%
\newcommand{\ie}{{\it i.e.}}%
\newcommand{\etal}{{\it et al.}}%
\newcommand{\etc}{{\it etc}}%

\newcommand{\aggregate}[2]{\underset{#2}{\operatornamewithlimits{#1\ }}}

\newcommand{\mcal}[1]{{\cal{#1}}}
\newcommand{\calA}{\mbox{${\cal A}$}}
\newcommand{\calB}{\mbox{${\cal B}$}}
\newcommand{\calC}{\mbox{${\cal C}$}}
\newcommand{\calD}{\mbox{${\cal D}$}}
\newcommand{\calE}{\mbox{${\cal E}$}}
\newcommand{\calF}{\mbox{${\cal F}$}}
\newcommand{\calG}{\mbox{${\cal G}$}}
\newcommand{\calH}{\mbox{${\cal H}$}}
\newcommand{\calI}{\mbox{${\cal I}$}}
\newcommand{\calJ}{\mbox{${\cal J}$}}
\newcommand{\calK}{\mbox{${\cal K}$}}
\newcommand{\calL}{\mbox{${\cal L}$}}
\newcommand{\calM}{\mbox{${\cal M}$}}
\newcommand{\calN}{\mbox{${\cal N}$}}
\newcommand{\calO}{\mbox{${\cal O}$}}
\newcommand{\calP}{\mbox{${\cal P}$}}
\newcommand{\calQ}{\mbox{${\cal Q}$}}
\newcommand{\calR}{\mbox{${\cal R}$}}
\newcommand{\calS}{\mbox{${\cal S}$}}
\newcommand{\calT}{\mbox{${\cal T}$}}
\newcommand{\calU}{\mbox{${\cal U}$}}
\newcommand{\calV}{\mbox{${\cal V}$}}
\newcommand{\calW}{\mbox{${\cal W}$}}
\newcommand{\calX}{\mbox{${\cal X}$}}
\newcommand{\calY}{\mbox{${\cal Y}$}}
\newcommand{\calZ}{\mbox{${\cal Z}$}}

\begin{abstract}
This paper introduces a simple yet effective data-centric approach for the task of improving persona-conditioned dialogue agents. Prior model-centric approaches unquestioningly depend on the raw crowdsourced benchmark datasets such as Persona-Chat. In contrast, we aim to fix annotation artifacts in benchmarking, which is orthogonally applicable to any dialogue model. Specifically, we augment relevant personas to improve dialogue dataset/agent, by leveraging the primal-dual structure of the two tasks, predicting dialogue responses and personas based on each other. Experiments on Persona-Chat show that our approach outperforms pre-trained LMs by an 11.7 point gain in terms of accuracy.
\end{abstract}

 \section{Introduction} \label{sec:intro}

In personalized dialogue agents, \emph{persona grounding} has been a long-standing goal to improve both human-likeness and dialogue consistency. For example, when given a profile description such as ``\textbf{\texttt{I am a doctor.}}'' and ``\textbf{\texttt{I don't eat meat}}'', dialogue agents aim to plausibly respond to dialogue context while endowing the machine with the persona, \eg, ``\emph{I am now working at hospital}'' and ``\emph{I went vegan.}'' respectively. To learn and evaluate such grounding, recent research~\cite{kim2020will,li2020aloha,song2020generating,zhang2019consistent} has proposed to encode personas based on pre-trained language models (PLMs)~\cite{devlin2018bert,liu2019roberta,radford2019language}, and many persona-conditioned dialogue benchmarks have been released such as Persona-Chat~\cite{zhang2018personalizing}, where crowdworkers role-play following the given description of personas to populate dialogues.

In spite of such recent significant progress, we argue that there is much room for improving persona-grounded dialogue agents in the data-centric view. Specifically, as crowdworkers often miss out on stating detailed experience and knowledge, it is reported that several linguistic biases exist in the persona-grounded dialogue datasets, for example, crowdworkers tend to reuse some terms of persona descriptions to script dialogue utterances, \eg, ``\emph{I am a doctor.}'', in which words trivially overlap to a persona sentence, or merely script utterances as a paraphrase, \eg, ``\emph{I work as a doctor.}'' or ``\emph{I don't eat any meat.}''.

\begin{figure}[t!]
    \centering
    \begin{tabular}{cc} 
        \multicolumn{2}{c}{\includegraphics[width=70mm]{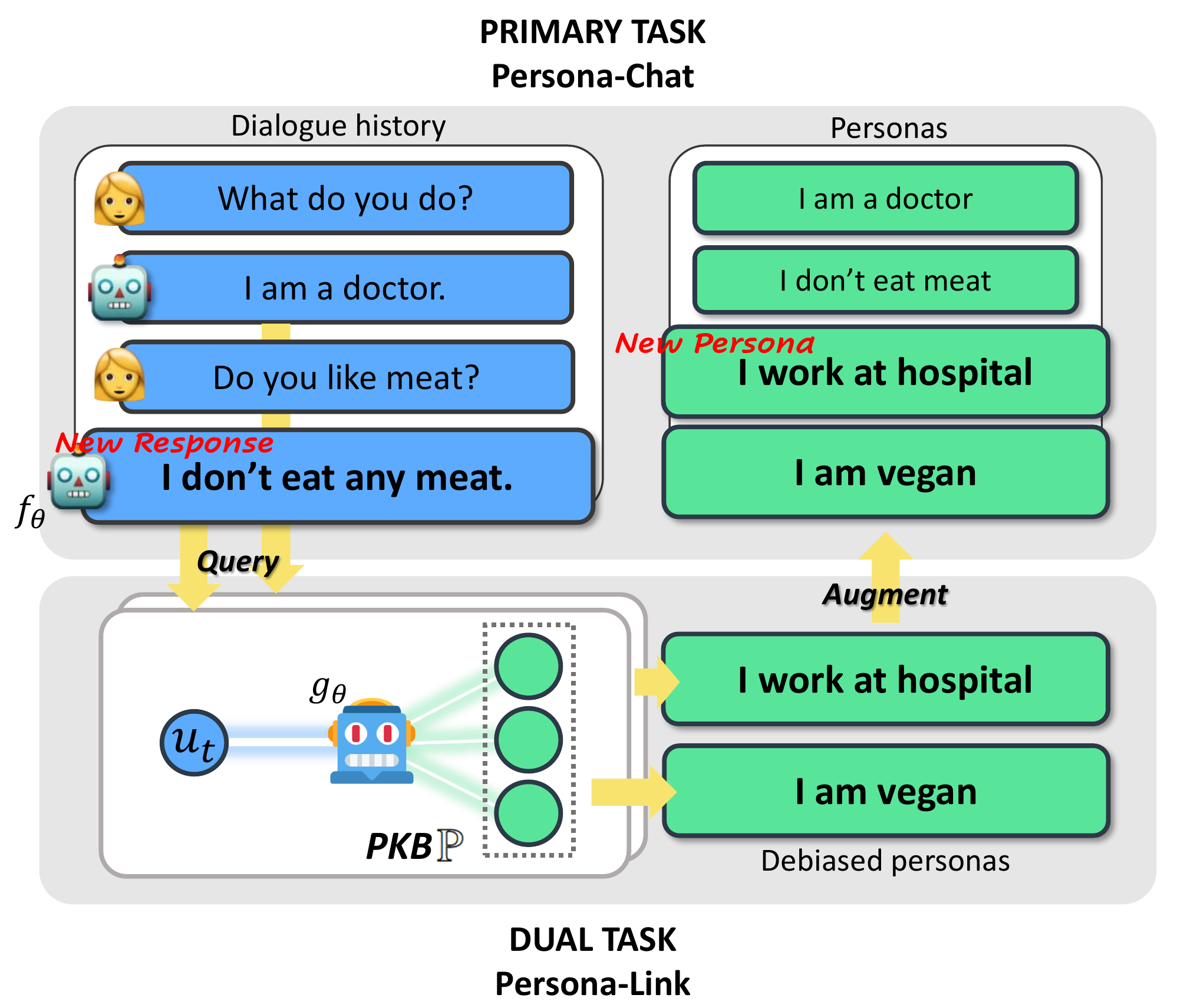}}\hfill
    \end{tabular}
    \caption{Illustration of the primal-dual task pipeline}\label{fig:personalink}
\end{figure}

Such annotation artifacts can be a bottleneck for learning to ground more engaging utterances and adversarially learning not to make contradictory utterances to personas. In practice, dialogue agents hardly respond in an engaging manner, \eg, ``\emph{I am now working at hospital.}'', loosening the linguistic ties from the given persona~\cite{ghazarian2021discol}. Also, on a challenging question such as  ``\emph{Are you vegan?}'' and ``\emph{Do you like steak?}'', dialogue agents may violate consistency~\cite{welleck2018dialogue}, \eg, responding ``\emph{No}'' and ``\emph{Yes}'' respectively. A straightforward way to mitigate these phenomena is to manually fix the dataset with additional annotation costs~\cite{bowman2021will}, by asking another crowdworkers to rewrite/augment the dialogue utterances, but it is too expensive on a large scale.

As motivated above, in this work, we propose to automatically improve the persona-conditioned dialogue dataset to learn better persona-grounded dialogue agents. As illustrated in Figure~\ref{fig:personalink}, our key idea is to infer and add new personas from a dialogue, \eg, ``\emph{I am a doctor.}'' $\mapsto$ \textbf{\texttt{I work at hospital}} or ``\emph{I don't eat any meat.}'' $\mapsto$  \textbf{\texttt{I am vegan}}. For that, we present a novel dual task framework, where the primal task Persona-Chat is learning persona-conditioned dialogue models (\ie, predicting utterances based on persona), whose personas are augmented by the dual yet secondary task, namely Persona-Link (\ie, predicting personas for utterances reversely). Our framework first learns the Persona-Link models, which use dialogue contexts in Persona-Chat as a query to augment its relevant personas, then learns the Persona-Chat models with the augmented personas for better persona grounding.

Straightforwardly, reversing the Persona-Chat dataset can provide the utterance-to-persona alignments as linking supervisions to train the Persona-Link models. However, using naive closed alignments from individual dialogue episodes may inherit the linguistic bias from Persona-Chat, which disqualifies the ultimate role of debiasing Persona-Chat. To tackle this challenge, we first adapt ideas from semantic matching~\cite{wu2020scalable} to our dual task, which enables augmenting open alignments out of individual dialogue episodes, \ie, less linguistically-biased alignments from the whole Persona-Chat corpus. Then, furthermore, we also leverage commonsense expansion~\cite{majumder2020like}, which enables augmenting more open alignments out of the Persona-Chat corpus itself, in a systematic way.

Our main contributions are summarized as follows:
(1) We propose an iterative framework of primal-dual tasks to debias the Persona-Chat dataset/model without any human effort. (2) We automate the learning of Persona-Link models from the Persona-Chat dataset, as a desired form of debiasing Persona-Chat. (3) Our extensive experiments validate that, along with linked personas, the response accuracy significantly increases by 11.7\% point on Persona-Chat compared to that of using the raw dataset.

\section{Primal-Dual Task Framework} \label{sec:task}

\subsection{Primal Task: \textsc{Persona-Chat}} \label{sec:task.1}
The goal of the Persona-Chat task~\cite{zhang2018personalizing} is to personalize dialogue agents by grounding their utterances to the given personas. 
The original dataset involves dialogues between pairs of speakers: each speaker is given a hypothetical profile, which is a few persona sentences that describe a character they will imitate, and is instructed to get to know the other. Formally, a profile is defined as a set of persona sentences $\calP=\{p_1,...,p_m\}$ and a dialogue is defined as a set of multi-turn utterances $\calU=\{u_1,...,u_n\}$, and a Persona-Chat dataset $\mathcal{D}_{\textsc{Chat}} = \{(\calP_i, \calU_i)\}_{i=1}^{N}$ where $N$ is the number of dialogues. Based on $\mathcal{D}_{\textsc{Chat}}$, when given persona sentences $\calP$, dialogue history $\calU$, and response space $\mathbb{U}$, the primal task aims at finding a model $f:(\calP,\calU)\mapsto \mathbb{U}$.
\begin{equation}\label{eq:f}
    f(\calP,\calU;\theta_{\textsc{Chat}}) \overset{\Delta}{=} \aggregate{argmax}{u \in \mathbb{U}}P(u|\calP,\calU;\theta_{\textsc{Chat}})
\end{equation}
where $\theta_{\textsc{Chat}}$ is the parameters to be learned for model $f_\theta$. In our setting, we adopt the response selection task~\cite{humeau2019poly}. As response space $\mathbb{U}$, the model has to pick the correct response from a set of 20 choices, where the remaining 19 were randomly chosen utterances from the evaluation set. Note that in a final system, however, one would retrieve from the entire training set of over 100k utterances, but this is avoided for speed reasons in common evaluation setups~\cite{wolf2019transfertransfo,humeau2019poly}.

\subsection{Dual Task: \textsc{Persona-Link}} \label{sec:task.2}
Following the primal-dual structure, the Persona-Link task can be defined as: given an arbitrary dialogue utterance $u$, the output of a linking model $g$ is its referent persona description in persona space $\mathbb{P}$. That is, the dual task aims at finding a linking model $g:u \mapsto \mathbb{P}$.
\begin{equation}\label{eq:g}
    g(u;\theta_{\textsc{Link}}) \overset{\Delta}{=} \aggregate{argmax}{p \in \mathbb{P}} P(p|u;\theta_{\textsc{Link}})
\end{equation}
where $\theta_{\textsc{Link}}$ is the parameters to be learned for model $g_\theta$. We formalize the dual task as a variant of entity linking system, where an utterance is linked to an entry in the PKB $\mathbb{P}$ of arbitrary size, being populated from the primal dataset $\mathcal{D}_{\textsc{Chat}}$, \ie, $\mathbb{P} = \bigcup_{i=1}^N \calP_i$. To learn model $g_\theta$, we adopt the same neural architecture of the primal task (\ie, Bi-encoder)~\cite{humeau2019poly} as a base model with the cross-entropy loss.\footnote{For more flexible application, inspired by~\cite{wu2020scalable}, we relax the collective linking in Eq. (\ref{eq:g}) into finer-level linking computing the utterance-to-persona score, \ie, $p(p_i|u_i;\theta_\textsc{Link})$.}


\begin{algorithm}[h!]
\fontsize{9.5}{11.5}\selectfont
\STATE{\textbf{Input:} Original data $\calD_{\textsc{Chat}}$}\\
\STATE{\textbf{Output:} Debiased data $\tilde{D}_{\textsc{Chat}}$, Debiased model $\tilde{\theta}_{\textsc{Chat}}$} \\
\STATE{$\calD_{\textsc{Link}}, \tilde{\calD}_{\textsc{Link}} \leftarrow \text{ReverseDataset}(\calD_{\textsc{Chat}})$~~~~~~}\tikzmark{top}\tikzmark{right}\\ 
\STATE{$\theta_{\textsc{Link}} \leftarrow \text{Train}(g_\theta, \calD_{\textsc{Link}})$}\\
\STATE{$\tilde{\theta}_{\textsc{Link}} \leftarrow \text{Train}(g_\theta, \tilde{\calD}_{\textsc{Link}}, \theta_{\textsc{Link}})$}\tikzmark{bottom}\\
\STATE{$\tilde{\calD}_{\textsc{Chat}} \leftarrow \text{AugmentPersona}(\calD_{\textsc{Chat}}, \tilde{\theta}_{\textsc{Link}})$}
\STATE{$\tilde{\theta}_{\textsc{Chat}} \leftarrow \text{Train}(f_\theta, \tilde{D}_{\textsc{Chat}})$}\tikzmark{toptop}\tikzmark{bbottom}\\
\STATE{\textbf{return}~ $\tilde{D}_{\textsc{Chat}}, \tilde{\theta}_{\textsc{Chat}}$}
\caption{Primal-Dual Task Framework}\label{alg:duallearning}
\end{algorithm}
\AddNote{top}{bottom}{right}{Persona-Link}
\AddNotee{toptop}{bbottom}{right}{Debiased Persona-Chat}

\begin{figure*}[t]
    \centering
    \begin{tabular}{c} 
        \includegraphics[width=17cm]{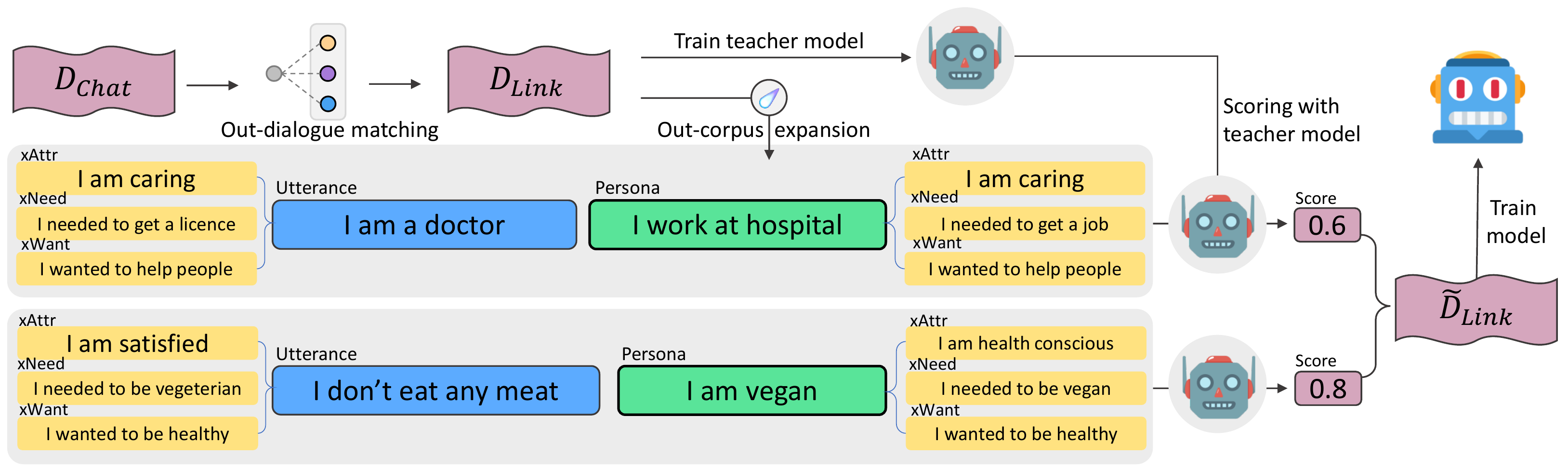}\hfill
    \end{tabular}
    \caption{Overall procedure of learning linking models at semantic and commonsense level for Persona-Link}\label{fig:overview}
\end{figure*}

\subsection{Primal-Dual Task Pipeline}\label{sec:task.3}
We introduce a new framework that exploits the duality~\cite{xia2017dual} of Persona-Chat and Persona-Link tasks where the input (utterance) and output (persona) of the dual task are roughly the inverse of its primal task.

In the training phase of the primal task, our ultimate goal is to transform the original dialogue dataset $\calD_{\textsc{Chat}}$ into the debiased dataset $\tilde{\calD}_{\textsc{Chat}}$ by augmenting new personas, then train the debiased dialogue model $f_\theta$ from $\tilde{\calD}_{\textsc{Chat}}$, which reduces the model dependence on linguistic bias between personas and utterances. Algorithm~\ref{alg:duallearning} describes its overall procedure. In the proposed framework, the linking model $g_\theta$ should be learned in advance to augment plausible personas per a given utterance. However, the key challenge is collecting the supervisory information as the training data of $g_\theta$ (denoted as $\calD_\textsc{Link}$ or $\tilde{\calD}_\textsc{Link}$), which is capable of injecting the debiasing ability to $g_\theta$. As illustrated in Figure~\ref{fig:overview}, we mainly discuss this in the subsequent section.

Once an arbitrary linking model $g_\theta$ is obtained as the desired debiasing function, we leverage $g_\theta$ also in the inference phase of the primal task, not only in training time. Specifically, as illustrated in Figure~\ref{fig:personalink}, a new persona is interactively augmented per new response to ground the dialogue agents to debiased persona-response alignments. This procedure is repeated until the dialogue ends.


\section{Reversing \textsc{Persona-Chat} to \textsc{Persona-Link}}\label{sec:data}

A straightforward way to learn the dual task, Persona-Link, is to reverse input and output sides of the primal task each other~\cite{wang2021effective,su2020dual,zhu2020dual}. However, in contrast to prior work, such a naive approach disqualifies the ultimate role of debiasing Persona-Chat since the Persona-Link data and model can inherit even the linguistic biases presented in the Persona-Chat dataset. Thus, the Persona-Link data should involve the desired asymmetric characteristics with Persona-Chat. We now present the three phases for learning Persona-Link models.

\subsection{Phase 1: Out-dialogue Semantic Matching}\label{sec:data.1}
Basically, as supervisory data for Persona-Link, we can collect the plausible alignments of persona and utterances in individual dialogue episodes of the Persona-Chat dataset, by using their semantic relationship. Specifically, we leverage natural language inference (NLI), a task of determining whether a hypothesis (\eg, persona) can be inferred from the given premise (\eg, utterance). The hypothesis sentence is classified into three categories: \texttt{Entailment} (true), \texttt{Contradiction} (false), and \texttt{Neutral} (undetermined). We adopt a RoBERTa model~\cite{liu2019roberta} trained on MNLI~\cite{wang2018glue}. By NLI, the minimal sampling to populate $\calD_{\textsc{Link}}$ is to consider the \emph{in-dialogue} matching of persona-utterance pair $(u,p)$ as candidates, where $u$ and $p$ are derived from the same dialogue episode. However, as mentioned earlier, such $(u,p)$ pairs in individual dialogue episodes reportedly suffer from trivial word overlaps.

In contrast, we perform the \emph{out-dialogue} matching of all possible combinations in $\calD_{Chat}$, which is stochastically less overlapped at lexical level. That is, such a simple strategy can contribute to mitigating the linguistic bias in the NLI-driven alignments, which we call seed Persona-Link dataset. 

\newtheorem{definition}{Definition}
\begin{definition}{(Seed Persona-Link Dataset)}\label{def:type1}
Given $\mathcal{D}_{\textsc{Chat}} = \{(\calP_j, \calU_j)\}_{j=1}^{N}$, a seed Persona-Link dataset can be defined as $\calD_{\textsc{Link}} = \{(u_i, p_i, y_i)\}_{i=1}^M$ where $u_i$ and $p_i$ are drawn from $\mathcal{D}_{\textsc{Chat}}$, \ie, $u_i \in \bigcup_{j=1}^N \calU_j$ and $p_i \in \bigcup_{j=1}^N \calP_j$, and binary label $y_i$ is captured by a NLI classifier. If $(u_i,p_i)$ is inferred as \texttt{Entailment}, $y_i=1$, and $y_i=0$ otherwise.
\end{definition}

\subsection{Phase 2: Out-corpus Commonsense Expansion}\label{sec:data.2}

Linguistic bias comes from the limited knowledge and experience of individual crowdworkers, which motivates the out-dialogue matching of personas and annotated utterances. Beyond the Persona-Chat corpus, if someone would have world-level knowledge, she may willingly annotate more engaging, less linguistically-biased utterances for a given persona. For example, utterance ``\emph{I don't eat any meat.}'' can involve commonsense attributes `\emph{I needed to be vegeterian}' or `\emph{I wanted to be healthy}', which enable drawing a new persona ``\texttt{I am vegan}'' from PKB, which are not captured by semantic matching (\ie, predicted as Neutral).

If a linking model is trained on sentence pairs of only similar semantics, such a potentially relevant mapping cannot be captured for generalization in inference time. Thus, since personas and utterances are instances of world events that often imply real-world consequences or richer information, we propose to exploit such commonsense attributes as ``anchors''. It may ensure the learned representations are well associated via reasoning effects~\cite{liu2020commonsense,majumder2020like} beyond the semantically close alignments.

Specifically, we populate an augmented dataset $\tilde{\calD}_{Link}$ that expands pairs in ${\calD}_{Link}$ based on commonsense knowledge. For commonsense expansion, we capture implicit attributes of either personas and utterances and annotate them as metadata, using GPT2 based commonsense knowledge generators~\cite{hwang2021comet} (Appendix A). The commonsense attributes are surrounded by special tokens and concatenated into a single sequence with persona or utterance. 

\begin{table*}[t!]
\begin{center}\small
\renewcommand\thetable{2}
{
\begin{tabular}{cccccc}
\hline
  \noalign{\hrule height0.8pt} 
   Dialogue model & Linking model & R@1/20 $\uparrow$ & R@5/20 $\uparrow$ & MRR $\uparrow$ & Contradict@1 $\downarrow$ 
  \TBstrut\\
  \hline
  \noalign{\hrule height0.8pt}
    Bi-encoder & \multirow{2}{*}{N/A} & 0.814 & 0.973 & 0.882 & 0.075  \TBstrut \\
    Cross-encoder &  & 0.884 & 0.991 & 0.930 & 0.040  \TBstrut \\ \hline
    \multirow{4}{*}{Bi-encoder} & Manual Paraphrasing & 0.758 & 0.961 & 0.876 & 0.088  \TBstrut \\
    &Paraphrasing & 0.769 & 0.963 & 0.851 & 0.094  \TBstrut \\
    &Bi-encoder & 0.881 & 0.987 & 0.927 & 0.053  \TBstrut \\
    &Cross-encoder & 0.881 & 0.988 & 0.928 & 0.043  \TBstrut \\ \hline
    \multirow{2}{*}{Bi-encoder} &Persona-Link (small PKB) & 0.904 & 0.992 & 0.943 & 0.039  \TBstrut \\
    &Persona-Link (large PKB) & \textbf{0.931} & \textbf{0.993} & \textbf{0.959} & \textbf{0.027}  \TBstrut \\
  \hline
  \noalign{\hrule height0.8pt} 
\end{tabular}
}
\end{center}
\caption{The performance of dialogue models on Persona-Chat testset}
\label{tab:persona_aug}
\end{table*}


\begin{definition}{(Commonsense Persona-Link Dataset)}\label{def:type2}
Given $\calD_{\textsc{Link}} = \{(u_i, p_i, y_i)\}_{i=1}^M$, its commonsense-expanded set is defined as $\Tilde{\calD}_{\textsc{Link}} = \{(\Tilde{u}_i, \Tilde{p}_i, y_i)\}_{i=1}^M$ where expanded samples $\Tilde{u}_i$ and $\Tilde{p}_i$ are obtained by a commonsense expansion function $\psi:(u_i \text{~or~} p_i)\rightarrow (\Tilde{u}_i \text{~or~} \Tilde{p}_i)$. This function provides tuples that belong to nine relation types spanning over cause-effect interrelations between events: \texttt{xAttr}, \texttt{xEffect}, \texttt{xIntent}, \texttt{xNeed}, \texttt{xReact}, \texttt{xWant}, \texttt{oEffect}, \texttt{oReact}, \texttt{oWant}— where a prefix `x' indicates an effect or cause on the person and `o' denotes the same on others. We present more details of the commonsense expansion function in Appendix A.
\end{definition}

\subsection{Phase 3: Learning with Label Regularization}  \label{sec:data.3}

As many commonsense attributes are commonly shared between utterance and persona, they are likely to be relevant to each other although the NLI-driven label indicates negative. On the other hand, not always such an assumption is true since generated commonsense attributes are often ambiguous or over-claimed. These concerns motivate us to better regularize learning on $\tilde{\calD}_{\textsc{Link}}$ by well-calibrated soft labels.
That is, the expanded set $\tilde{\calD}_{\textsc{Link}}$ on the denser space may not strictly follow the pre-annotated binary labels from NLI.

To address this, we argue that the linking model trained on $\calD_{\textsc{Link}}$ can be a good reference point. Specifically, we first train a linking model $\theta_{\textsc{Link}}$ only with the semantic-level Persona-Link dataset $\calD_{\textsc{Link}}$, which can further perform inference over new data samples $(\tilde{u}, \tilde{p})$ to compute their outputs as new labels $P(\tilde{p}|\tilde{u};\theta_{\textsc{Link}})$. By regularizing by the semantic feature space of $\theta_{\textsc{Link}}$, the commonsense-expanded dataset can have well-calibrated labels for encoding commonsense attributes. We thus train another linking model $\tilde\theta_{\textsc{Link}}$ with dual goals of following not only the original hard labels but also new soft labels as:
\begin{align} 
    \theta_{\textsc{Link}} = \text{argmin}_{\theta}  &\sum_{(u,p,y) \in \mathcal{D}_{\textsc{Link}}}  \calL( y, P(p|u;\theta))\\
    \tilde{\theta}_{\textsc{Link}} = \text{argmin}_{\theta}  &\sum_{(\tilde{u},\tilde{p},y) \in \tilde{\mathcal{D}}_{\textsc{Link}}}  \calL( y, P(\tilde{p}|\tilde{u};\theta))\\
    &+ \lambda \cdot \calL( P(\tilde{p}|\tilde{u};\theta_{\textsc{Link}})  , P(\tilde{p}|\tilde{u};\theta))\notag
\end{align}
where $\lambda$ is a preference weight with the distillation loss. To compute the linking score $P(p|u;\theta)$ and the cross-entropy loss $\calL$, based on the Bi-encoder architecture allowing for fast and real-time inference, we follow the optimization procedure~\cite{logeswaran2019zero,humeau2019poly,jeong2021label} of retrieval-based models (\eg, information retrieval, entity linking, and response selection): The network is trained to maximize the score of the correct persona $p$ with respect to the (randomly sampled) personas of the same batch (\ie, \emph{in-batch negatives}).

\subsubsection{Inference}
Once a linking model $\tilde{\theta}_{\textsc{Link}}$ is learned, given a utterance $u$ in Persona-Chat, we follow Eq. (\ref{eq:g}) to augment a new persona into the original dialogue episode. Considering that the development/test sets of Persona-Chat are unseen, as PKB $\mathbb{P}$, we only use a list of personas involved in the training set of Persona-Chat at least once.


\begin{table*}[t!]
\begin{center}\small
\renewcommand\thetable{2}
{
\begin{tabular}{llccccccccc}
\noalign{\hrule height0.8pt}
  \multicolumn{1}{c}{\multirow{3}{*}{Linking model}} & & \multicolumn{9}{c}{Dialogue model}
    \\   &
  &
  \multicolumn{4}{c}{Pretrained on Wikipedia} &
  \multicolumn{1}{l}{} &
  \multicolumn{4}{c}{Pretrained on Reddit}
  \\ \cline{3-6} \cline{8-11}
\multicolumn{1}{c}{} &
  &
  R@1/20$\uparrow$ &
  R@5/20 $\uparrow$ &
  MRR $\uparrow$ &
  Ctrd.@1 $\downarrow$ &
  &
  R@1/20$\uparrow$ &
  R@5/20 $\uparrow$ &
  MRR $\uparrow$ &
  Ctrd.@1 $\downarrow$ 
  \\
\noalign{\hrule height0.8pt} 
\multicolumn{11}{l}{\textbf{No persona} (0\%)}         \\
N/A        &  & 0.591 & 0.870 & 0.711 & 0.174 &  & 0.659 & 0.900 & 0.766 & 0.144   \\
Bi-encoder &  & 0.699 & 0.936 & 0.802 & 0.120 &  & 0.773 & 0.960 & 0.853 & 0.094   \\
Persona-Link &  & \textbf{0.733} & \textbf{0.956} & \textbf{0.828} & \textbf{0.113} &  & \textbf{0.805} & \textbf{0.970} & \textbf{0.876} & \textbf{0.078}   \\ \hline
\multicolumn{11}{l}{\textbf{Incomplete persona} (80\%)}          \\
N/A        &  & 0.768 & 0.953 & 0.848 & 0.094 &  & 0.825 & 0.973 & 0.889 & 0.073   \\
Bi-encoder &  & 0.768 & 0.966 & 0.850 & 0.094 &  & 0.845 & 0.984 & 0.905 & 0.064   \\
Persona-Link &  & \textbf{0.785} & \textbf{0.972} & \textbf{0.864} & \textbf{0.093} &  & \textbf{0.858} & \textbf{0.985} & \textbf{0.913} & \textbf{0.058}  \\ \hline
\multicolumn{11}{l}{\textbf{Original persona} (100\%)}           \\
N/A        &  & 0.814 & 0.973 & 0.882 & 0.075 &  & 0.868 & 0.986 & 0.919 & 0.054  \\

\noalign{\hrule height0.8pt} 
\end{tabular}
}
\end{center}
\caption{The response quality of Persona-Chat models, observing debiased personas added at test time. In advance, in-dialogue persona sentences were randomly removed, and link models augmented in-context personas based on dialogue histories.}
\label{tab:chat_performance}
\end{table*}


\section{\textsc{Persona-Chat} Evaluation}\label{sec:exp.pc}

As bias analysis in benchmark datasets is a non-trivial problem \cite{bowman2021will,torralba2011unbiased}, we measure the response quality of dialogue models as a proxy of bias mitigation. We measure Recall@$k/N$ and MRR as the model performance to the gold utterance, where $N=20$ and $k=[1, 5]$. Another metric is Contradict@1, indicating the textual disagreement judged by an NLI model: the proportion of contradictory responses in the top-1 candidates returned by dialogue agents, \ie, consistency error ratio.

\subsection{Performance with White-box Training} \label{sec:exp.pc.1}

For the experiment on white-box settings (\ie, augmenting personas in both training and inference time), we consider following six models including two Persona-Link variants as the candidates to be analyzed.

1) \textbf{Paraphrasing}: As unsupervised data augmentation~\cite{xie2019unsupervised}, we consider an off-the-shelf paraphrasing system in \cite{mallinson2017paraphrasing}, where personas were translated into a foreign language and back-translated as paraphrases.

2) \textbf{Manual Paraphrasing}: As labor-intensive augmentation, we use manually revised persona sentences additionally presented in Persona-Chat, where extra workers rephrased them to remove trivial word overlap.

3) \textbf{Bi-encoder}: As an IR baseline of linking model~\cite{wu2020scalable} trained on $\calD_{\textsc{Link}}$, we use two transformers that separately encode utterance and persona, and multi-sentence scoring over the pre-computed cache of personas.

4) \textbf{Cross-encoder}: We consider using a single transformer~\cite{humeau2019poly} that uses a richer self-attention mechanism, jointly encoding utterance and persona in $\calD_{\textsc{Link}}$ to obtain the final representation, which is impractical for real-time use.

5) \textbf{Persona-Link (small PKB)}: Our proposed linking model with a smaller Persona Knowledge Base (PKB), caching 1K personas which are randomly sampled from the primal dataset $\mathcal{D}_{\textsc{Chat}}$.


6) \textbf{Persona-Link (large PKB)}: Using the same model, we present the original version of our approach, where all 5K personas from $\mathcal{D}_{\textsc{Chat}}$ are plugged into its PKB.

Table \ref{tab:persona_aug} shows the overall performance of dialogue models trained with different linking models. Models paired with Persona-Link variants significantly outperform other baselines for all measures. This suggests that training dialogue model with well augmented personas helps the model to be more persona-grounded and consistent which is a primary goal in persona-based dialogue literature. Furthermore, as the performance gain increases with the size of PKB used for Persona-Link, we believe that populating bigger PKB improves the performance.
Another observation is that Persona-Link variants outperform both manual and automatic paraphrasing approaches on every measure. Even though paraphrasing techniques might contribute to richer representation of personas, we argue that the paraphrasing approaches are limited to mitigating the trivial word overlap while utterances can be expressed in much flexible way in persona-based dialogue. 
While Bi- and Cross-encoder linking models result in better performance compare to the paraphrasing-based linking models due to out-dialogue matching which aims to mitigate linguistic bias, Persona-Link variants outperform in both measures. Especially, we observe all Persona-Link variants outperform Cross-encoder which compromises high computational cost for performance. We further analyze how Persona-Link achieves better performance in Persona-Link Evaluation section.

\subsection{Performance with Black-box Training} \label{sec:exp.pc.2}

To demonstrate that the augmented persona is not over-claimed by out-dialogue matching or out-corpus expansion, we compare the quality of augmented persona using dual task models. For that, we simulate a setting where none or few persona information is provided and new personas are augmented only in inference time by using off-the-shelf dialogue agents (\ie, learned without augmenting personas in training time). Inspired by the evaluation method of measuring the relevance of generated response in \cite{su2020diversifying}, we use the pre-trained dialogue agents as our diagnosis model assuming that the response quality of the model reflects the suitability of augmented persona in given dialogue contexts. That is, the suitability of the augmented persona to the context would be low if the augmented personas, which replace partial or entire persona of dataset, cause the failure of pre-trained dialogue agents.
For the diagnosis agents, we adopt two dialogue models of Bi-encoder; each pretrained on Wikipedia (BERT-like; \citealt{devlin2018bert}) or Reddit (an adapted setup for dialogue; \citealt{mazare2018training}), and both fine-tuned on $\calD_{\textsc{Chat}}$. Note that, during fine-tuning, the dialogue agents only observe the original personas. As dual task models, we adopt two linking models; Bi-encoder trained on $\calD_{\textsc{Link}}$ and our Persona-Link trained on $\tilde{\calD}_{\textsc{Link}}$.

Specifically, the original persona sentences were removed at a given percentage; \textbf{No persona} eradicates the whole persona information and \textbf{Incomplete persona} discards 20\%. In test time, linking models generate personas based on the utterance histories and merge them into the sample for testing dialogue models\footnote{For fair comparison, all utterances on the development set remained unseen by linking models. Also, for faster computation, all utterances in dialogue were concurrently processed in batch.}. With persona augmentation, Table~\ref{tab:chat_performance} shows the overall performance of dialogue models. In each setting, we report the base performance with information limit, where no linking model is available, \ie, N/A. In the last row, we report the upper bound performance, where dialogue models access full information of \textbf{Original persona}.

First, compared to N/A, we observe linking models significantly improve (+14.6\% at most) response quality by backtracking personas from utterances for the augmentation. Dialogue models seem to benefit from augmented personas, even though the models haven't seen them in training process, suggesting that personas provided by Persona-Link successfully keep contextual relevance to the given dialogues. Second, Persona-Link models outperform Bi-encoder models in all settings. As the only difference between the two is commonsense expansion with regularized training, our approach seems to enrich dual task with common knowledge beyond the alignments in Persona-Chat. Thirdly, the performance gain is the greatest in the persona-free setup, which is common in real-world scenarios. This suggests that our persona augmentation may boost dialogue agents in bootstrapping user profiles on-the-fly. As observing only debiased personas improves a trained agent, we conclude that our approach efficiently counteracts linguistic bias in Persona-Chat.

Now, we examine how persona augmentation calibrates model prediction in response retrieval, by showing an example in Table~\ref{tab:chat_case}. After removing a persona in advance, a linking model augments its own personas into context based on dialogue history. Observing additional information relevant to utterances (\ie meal) and linguistically debiased (\ie steak), the model prediction was calibrated online to retrieve the gold response, which is more consistent and engaging. Note that we use the same dialogue model to retrieve both responses. (See Appendix C for details).

\begin{table}[t!]
\begin{center}\small
\renewcommand\thetable{2}
{
\begin{tabular}{l}
\hline
  \noalign{\hrule height0.8pt}
    \textbf{Agent's personas:} \\
    My mom is a secretary. \\
    I am a bodybuilder. \\
    I have one brother. \\
    (-) I like to eat a lot of meat. \\
    (+) I am a meat eater. \\
    (+) My favorite meal is steak. \\
    \hline
    \noalign{\hrule height0.4pt}
    \textbf{Dialogue history:} \\
     \emph{User}: Hi there, how are you tonight? \\
     \emph{Agent}: Great. I just finished a huge steak. How are you? \\
     \emph{User}: I am good, drinking some scotch. \\
     \hline
    \textbf{Response without augmented personas:} \\
    \emph{Agent}: Cool! What are your hobbies?\\ 
    \hline
    \textbf{Response with augmented personas:}\\
    \emph{Agent}: I am major meat eater to build muscles.\\ 
  \hline
  \noalign{\hrule height0.8pt}
\end{tabular}
}
\end{center}
\caption{The model calibration in response retrieval task based on online persona augmentation. After removing persona (-), a linking model augmented personas  (+), which calibrates the prediction of dialogue agent, with any parameter updates.}
\label{tab:chat_case}
\end{table}

\section{\textsc{Persona-Link} Evaluation} \label{sec:exp.pl}

In this section, we conduct additional experiments to evaluate the proposed process for the dual task.

\subsection{Overall Linking Performance} \label{exp.pl.1}

To study the model accuracy to the dual task, we perform comparative experiments of Persona-Link. To find ground truth pairs for Persona-Link, we manually collect 300 utterance-persona pairs from the test data in Persona-Chat, including 230 unique ground truth personas with seven annotators, which achieve a very good agreement (Cohen's kappa = 0.86). The number of utterances per persona is 1.16. As in Persona-Chat evaluation, we measure Recall@$k$ and MRR as the model performance to the gold persona. 

We consider five IR baselines trained on $\mathcal{D}_{\textsc{Link}}$: 
\textbf{Cosine Similarity}, \textbf{BM25}~\cite{robertson1995okapi}, \textbf{K-NRM}~\cite{Xiong_2017} \textbf{Conv-KNRM}~\cite{dai2018convolutional}, and \textbf{Bi-encoder}~\cite{humeau2019poly}. Also, to investigate the impact of commonsense expansion and label regularization, we consider two variants of our approach using alternative expansion methods:
\textbf{EDA}~\cite{wei2019eda}, and \textbf{Paraphrasing}~\cite{mallinson2017paraphrasing}. More details for these baselines are presented in Appendix B.

\begin{table}[t!]
\begin{center}\small
\renewcommand\thetable{2}
{
\begin{tabular}{lcccc }
\hline
  \noalign{\hrule height0.8pt} 
   Linking Model & R@1 & R@10 & MRR 
  \TBstrut\\
  \hline
  \noalign{\hrule height0.8pt}
  Cosine Similarity & 0.108 & 0.349 & 0.190 \TBstrut \\
  BM25 & 0.491  & 0.792 & 0.595 \TBstrut \\
  K-NRM & 0.294  & 0.591 & 0.384  \TBstrut \\
  CONV-KNRM & 0.345  & 0.628 & 0.439 \TBstrut \\
  Bi-encoder & 0.583 & 0.871 & 0.684 \TBstrut \\
  \hline
  Ours (EDA) & 0.599  & 0.885 & 0.709 \TBstrut \\
  Ours (Para) & 0.610  & 0.885 & 0.715 \TBstrut \\
  Ours & \textbf{0.669}  & \textbf{0.922} & \textbf{0.759} \TBstrut \\
  \hline
  \noalign{\hrule height0.8pt} 
\end{tabular}}
\end{center}
\caption{Linking performance on Persona-Link}
\label{tab:performance}
\end{table}

\begin{figure}[t!]
     \centering
     \begin{subfigure}[t]{0.22\textwidth}
        \captionsetup{justification=centering}
         \includegraphics[width=\textwidth]{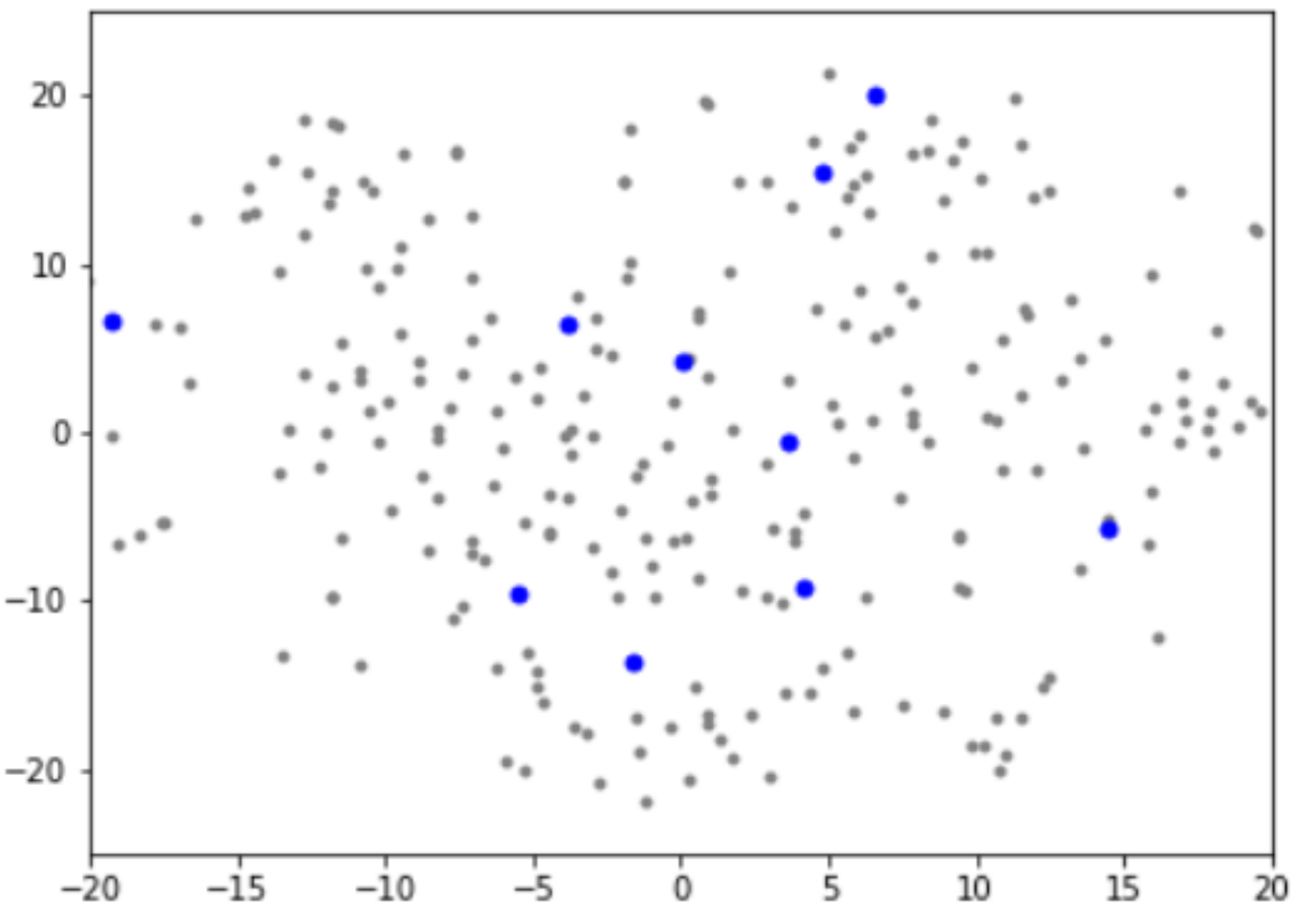}
         \caption{w/o commonsense expansion}
         \label{fig:visualization_before}
     \end{subfigure}
     \hfill
     \begin{subfigure}[t]{0.22\textwidth}
        \captionsetup{justification=centering}
         \includegraphics[width=\textwidth]{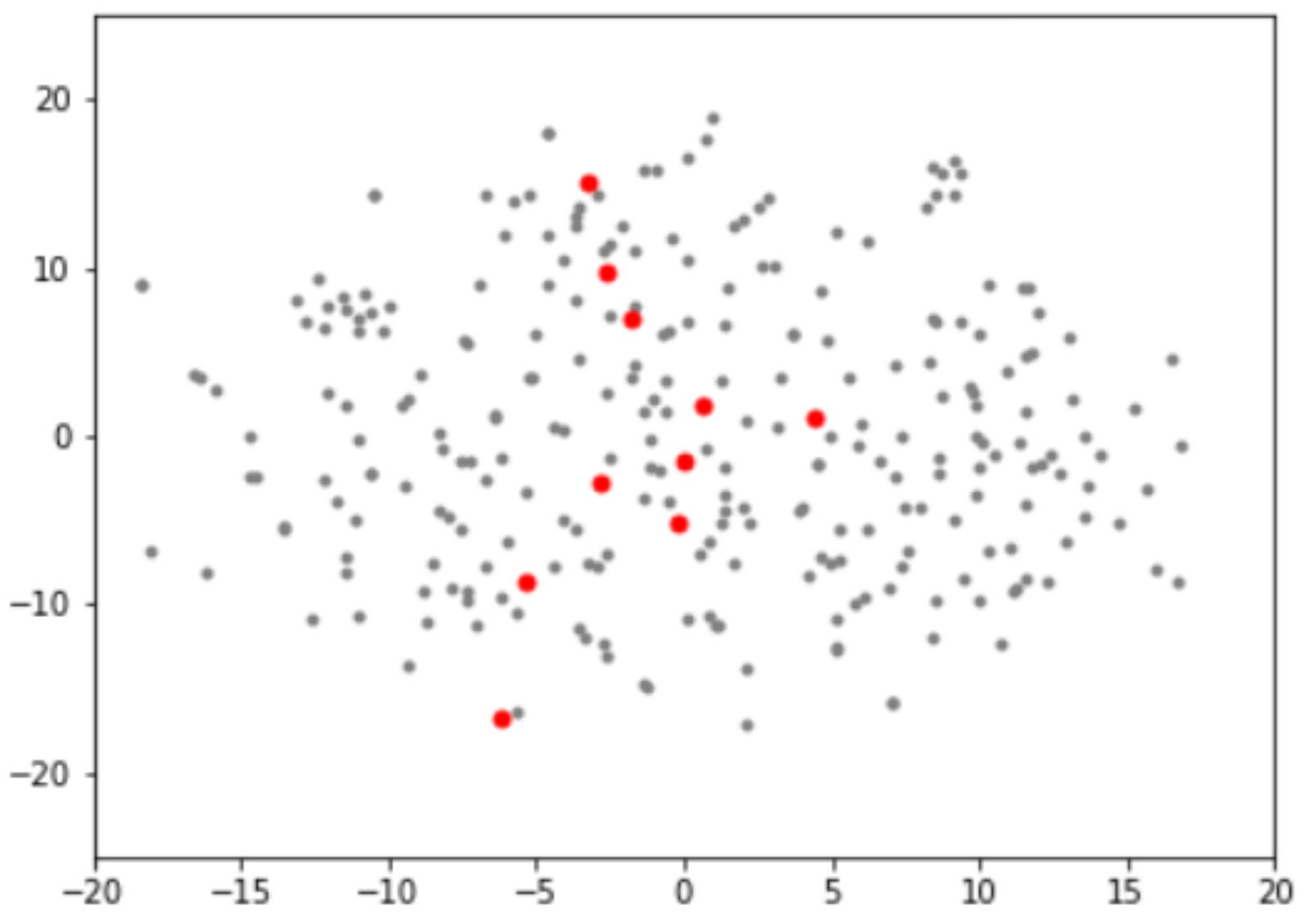}
         \caption{w/ commonsense expansion}
         \label{fig:visualization_after}
     \end{subfigure}
\caption{t-SNE visualization on the development set. The colored points indicate the utterance embeddings relevant to persona ``\textbf\texttt{{I have a marketing job}}''.}
\label{fig:visualization}
\end{figure}

\begin{figure}[t!]
    \centering
    \begin{tabular}{cc} 
        \multicolumn{2}{c}{\includegraphics[width=80mm]{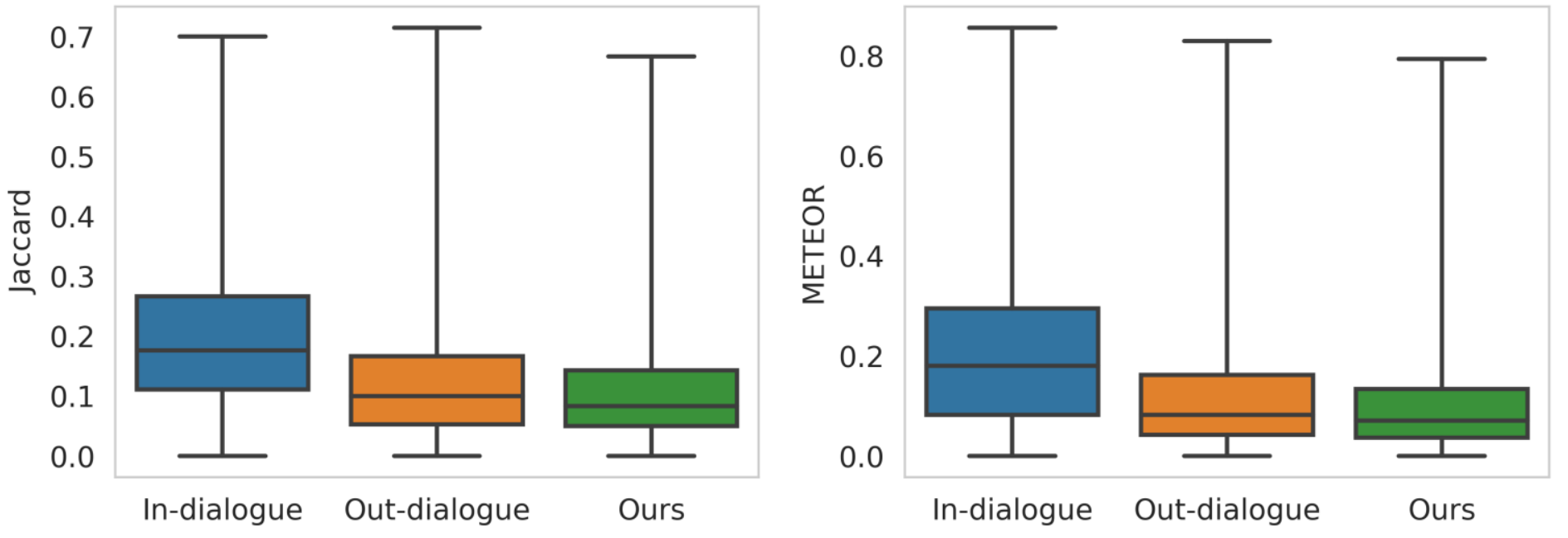}}\hfill
    \end{tabular}
    \caption{Similarity with respect to linking method. 
    }
    \label{fig:similarity}
\end{figure}

Table~\ref{tab:performance} shows the overall linking performance. 
First, we observe that among IR baselines trained on $\mathcal{D}_{\textsc{Link}}$, Bi-encoder achieves the best performance, which validates our choice of base model for the dual task. Second, all models trained on $\mathcal{D}_{\textsc{Link}}$ show poorer results than Ours and its variants using expansion methods with regularized training. This means that leveraging expanded dataset and calibrating labels into soft labels is more appropriate than naive IR models. Finally, Ours outperforms its variants (and other linking models as well) in all metrics. We note that while utterance or persona remains semantically similar by EDA or paraphrasing, Ours expands some commonsense attributes so that potentially relevant mapping can be captured. This supports the efficacy of exploiting commonsense as anchors, which may associates learned representations beyond the semantically close alignments. In Figure~\ref{fig:visualization}, we visualize the utterance embeddings without (Figure~\ref{fig:visualization}a) and with (Figure~\ref{fig:visualization}b) commonsense expansion. We observe that the utterance embeddings relevant to a persona gets closer with commonsense expansion, which can be a possible explanation for the performance boost of Ours exploiting commonsense over its variants.

\begin{figure}[t!]
    \centering
    \begin{tabular}{cc} 
        \multicolumn{2}{c}{\includegraphics[width=71mm]{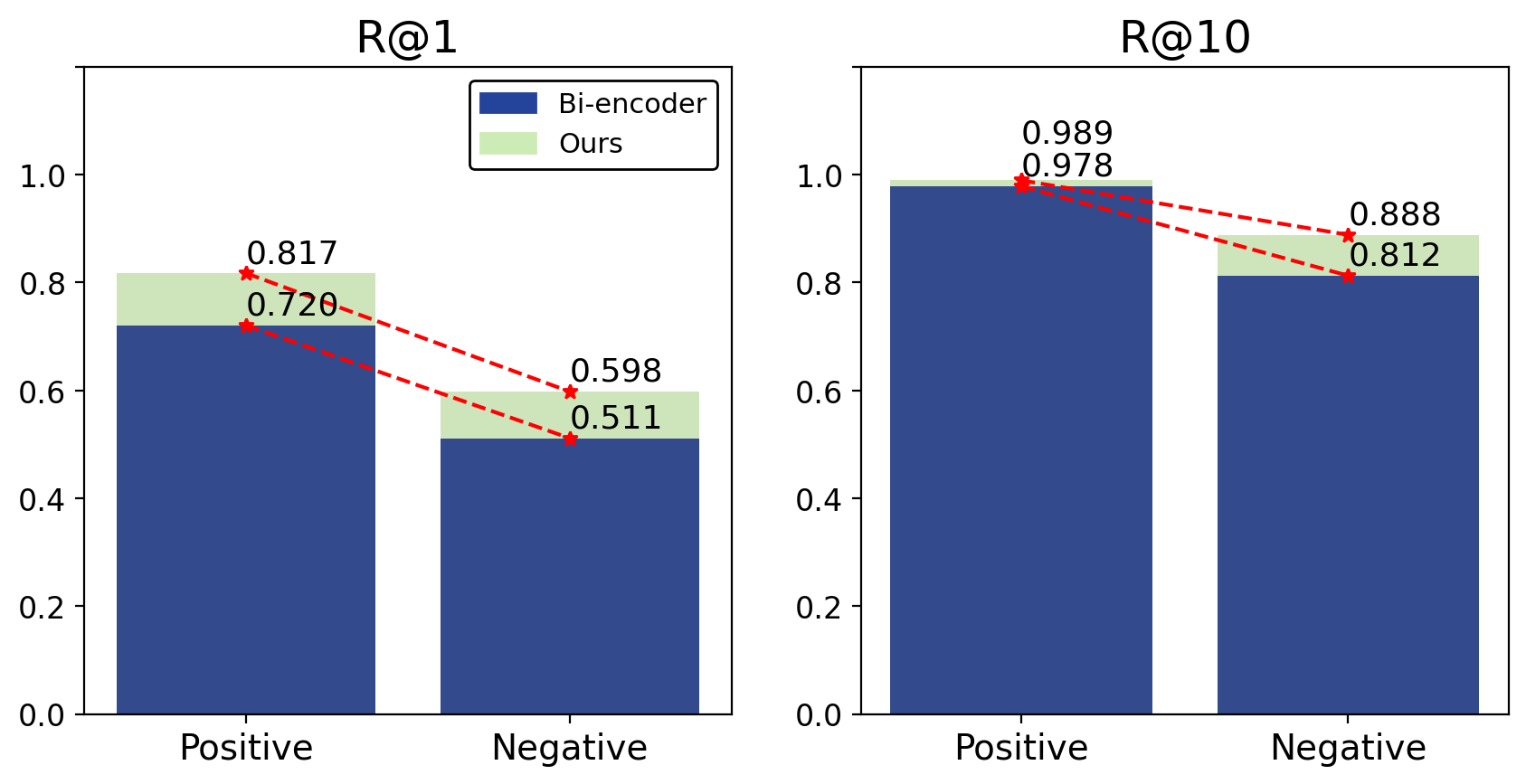}}\hfill
    \end{tabular}
    \caption{Performance with respect to PI result.}\label{fig:beforeafter2}
\end{figure} 

\begin{figure}[t!]
    \centering
    \begin{tabular}{cc} 
        \multicolumn{2}{c}{\includegraphics[width=71mm]{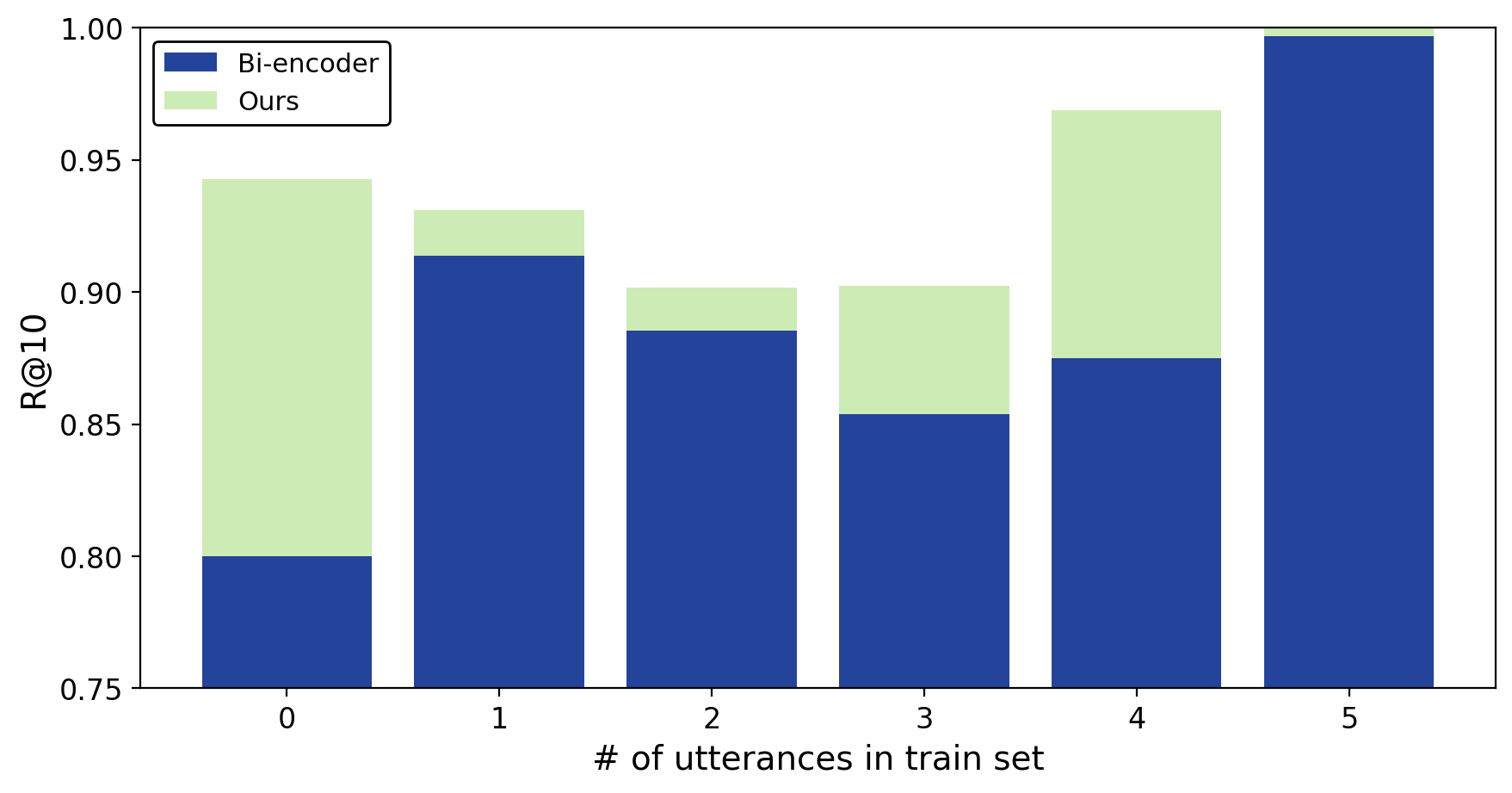}}\hfill
    \end{tabular}
    \caption{Performance with respect to data size.}\label{fig:beforeafter1}
\end{figure}

\subsection{Analysis on Linguistic Bias} \label{exp.pl.2}

In addition to the linking performance, we further analyze the linked utterance-to-persona alignments by linking models to see their linguistic bias. We plot the linguistic similarity between utterance-to-persona alignments in Figure~\ref{fig:similarity}. Following~\cite{park2019paraphrase}, we calculate Jaccard similarity and METEOR, widely known metric for lexical similarity and semantic similarity, respectively. Jaccard similarity measures word co-occurrence and METEOR measures exact, stem, synonym, and paraphrase matches.
For both metrics, an alignment with lower similarity can be seen as the one with less linguistic bias. Two models drop the original similarity from 0.2 to 0.1 on average, indicating the efficacy of out-dialogue semantic matching. Ours shows a slight decrease compared to out-dialogue model which confirms that Ours successfully leverages the commonsense expansion to align utterances to debiased persona.

We compare the performance of $\theta_{\textsc{Link}}$ and $\tilde{\theta}_{\textsc{Link}}$ in semantically different pair sets to investigate whether commonsense expansion helps the model link to debiased persona.
Here, we use a BERT model fine-tuned with a popular paraphrase identification (PI) dataset called Quora Question Pairs to split our utterance-persona pairs into semantically equivalent and non-equivalent sets. With this model, only 35.5\% of utterance-persona pairs are inferred as positive (\ie, paraphrase). 
For analysis, in Figure~\ref{fig:beforeafter2}, we split the overall utterance-persona pairs into positive and negative sets by the PI model, and ablate the performance of linking model with or without using commonsense knowledge and label regularization. 
As a result, while $\tilde{\theta}_{\textsc{Link}}$ shows better performance than $\theta_{\textsc{Link}}$ in all sets, $\tilde{\theta}_{\textsc{Link}}$ achieves much better performance in negative set of R@10. This demonstrates the advantage of using commonsense knowledge, which gives the model the ability to link semantically different pairs that leads $\tilde{\theta}_{\textsc{Link}}$ to link utterance to debiased persona.

Next, we evaluate how commonsense expansion affects when only a limited number of examples are provided in train time. Figure~\ref{fig:beforeafter1} shows the linking performance grouped by frequency of utterance-persona pairs in the training data in terms of R@10. Here, the performance means how well the model links utterances to their gold persona when the utterance-persona pair of the gold persona is not provided or few in train time. We observe that $\tilde{\theta}_{\textsc{Link}}$ has more balanced performance for any persona groups, which is the effect of using commonsense knowledge, which can work as pivots when the utterance and persona has similar commonsense in common. 
Especially, $\tilde{\theta}_{\textsc{Link}}$ shows the significant improvement for unseen personas (w/o any aligned training data) compared to $\theta_{\textsc{Link}}$, which shows that $\tilde{\theta}_{\textsc{Link}}$ may be applicable to zero-shot linking. Since $\tilde{\theta}_{\textsc{Link}}$ shows great performance to unseen utterance-persona pairs, we suggest to use our linking model with a large set of personas.

\section{Related Work} \label{sec:related}

Building personalized dialogue agents has been a popular task recently. \citet{zhang2018personalizing} introduced the Persona-Chat dataset, a crowd-sourced conversation dataset with persona information, to improve model engagingness and consistency. Based on such data, recent works focus on improving persona-grounded dialogue agent's performance~\cite{kim2020will,li2020aloha,song2020generating,huang2020challenges,zhang2019consistent,luo2019learning}. \citet{yavuz2019deepcopy} designed the DeepCopy model, which leverages copy mechanism to incorporate persona texts. \citet{song2019exploiting} integrated persona texts into the Per-CVAE model for generating diverse responses. However, prior work reports several linguistic bias which hamper learning to ground more engaging and consistent utterances.

Few recent works focused on augmenting grounding with commonsense knowledge with successful applications in open-domain dialogue agent~\cite{ghazvininejad2018knowledge,moon2019opendialkg}. \citet{ghazvininejad2018knowledge} generalizes the Seq2Seq approach to neural conversation models by combining conversational and KG data. \citet{moon2019opendialkg} propose DialKG walker model that learns the transitions of dialogue contexts as structured traversals over KG. 
In this work, we extend this effort into improving persona-grounded dialogue dataset in data-centric view via augmenting relevant personas.


\section{Conclusion} \label{sec:conclusion}
In this work, we propose a primal-dual task framework to de-bias the Persona-Chat dataset and its dialogue model without any human effort. By exploiting external LMs and knowledge distillation in a systematic way of overcoming the linguistic bias in Persona-Chat, our linking model was effective in improving 11.7\% in dialogue accuracy from BERT Bi-encoder model using the original Persona-Chat dataset. We hope future research
to leverage the graph structure of PKB with commonsense attributes for better persona linking.

\section{Acknowledgements} \label{sec:acknowledgements}
This work was partly supported by Institute of Information \& communications Technology Planning \& Evaluation (IITP) grant funded by the Korea government (MSIT) (No. 2020-0-01361, Artificial Intelligence Graduate School Program (Yonsei University)) and the National Research Foundation of Korea (NRF) grant funded by the Korea government (MSIT) (No. 2020-11-0863). Jinyoung Yeo is a corresponding author.

\bibliography{aaai22.bib}


\clearpage
\appendix

\section{Appendix A: Implementation Details}
\label{sec:appendix.1}
\subsection{COMET}
We use COMET~\cite{bosselut2019comet} which generates rich and diverse commonsense expansions of a given world event. 

It is a fine-tuned version of a pre-trained GPT2~\cite{radford2018improving} model on a pre-existing commonsense knowledge graph such as ATOMIC~\cite{sap2019atomic} that can generate novel nodes (events) and edges (relations). Specifically, ATOMIC provides tuples that belong to nine relation types spanning over cause-effect interrelations between events: \texttt{xAttr}, \texttt{xEffect}, \texttt{xIntent}, \texttt{xNeed}, \texttt{xReact}, \texttt{xWant}, \texttt{oEffect}, \texttt{oReact}, \texttt{oWant}— where a prefix `x' indicates an effect or cause on the person and `o' denotes the same on others. 
While we tried COMET finetuned on commonsense other than ATOMIC, such as ConceptNet, we found expansions unsatisfactory and do not report them.\footnote{Failure case includes that event-like persona (`\emph{I joined the air force, ... I learned how to fly!}') fails to match concept `\emph{fly}'.} For more details on COMET and ATOMIC, we refer the reader to~\cite{bosselut2019comet} and~\cite{sap2019atomic} respectively.

We use the COMET framework to generate expansions for not only persona but also utterance along the multiple relation types that ATOMIC provides (Figure~\ref{fig:overview}). Here we use only 6 ATOMIC relations with the prefix `x', since Persona-Link targets the utterances that the agent is talking about itself. We obtain different samples while decoding via greedy search from COMET for more accurate expansions (sacrificing diversity compared to beam search). 

For COMET expansions, we use the code\footnote{\url{https://github.com/atcbosselut/comet-commonsense}} released by the authors of COMET. We follow the default setting for this code to generate commonsense knowledge.

\subsection{MNLI model}
As a MNLI model, we use the public HuggingFace implementation of the RoBERTa-large model~\cite{liu2019roberta} fine-tuned with the Multi-Genre NLI dataset~\cite{wang2018glue}.

\begin{table}[t!]
\begin{center}\small
\renewcommand\thetable{2}
{
\begin{tabular}{l}
\hline
  \noalign{\hrule height0.8pt}
    \textbf{Personas} \\ 
    \hline
    I like to remodel homes. \\
    I like to go hunting. \\ 
    I like to shoot a bow. \\ 
    My favorite holiday is Halloween. \\
    \hline
    \noalign{\hrule height0.4pt}
    \textbf{Dialogue history} \\
    \hline
     \emph{Person1}: Hi, how are you doing? I am getting ready to \\
                ~~~~~~~~~~~~~~~ do some cheetah chasing to stay in shape. \\ 
     \emph{Person2}: You must be very fast.\\
                ~~~~~~~~~~~~~~~ Hunting is one of my favorite hobbies. \\
     \emph{Person1}: I am! For my hobby \\
                ~~~~~~~~~~~~~~~ I like to do canning or some whittling. \\
     \emph{Person2}: I also remodel homes \\
                ~~~~~~~~~~~~~~~ when I am not out bow hunting. \\
     \emph{Person1}: That is neat. When I was in high school \\
                ~~~~~~~~~~~~~~~ I placed 6th in 100m dash! \\
     \emph{Person2}: That is awesome. \\
                ~~~~~~~~~~~~~~~ Do you have a favorite season or time of year? \\
     \emph{Person1}: I do not. But I do have a favorite meat \\
                ~~~~~~~~~~~~~~~ since that is all I eat exclusively. \\
     \emph{Person2}: What is your favorite meat to eat? \\
     \emph{Person1}: I would have to say its prime rib. \\
                ~~~~~~~~~~~~~~~ Do you have any favorite foods? \\
     \emph{Person2}: I like chicken or macaroni and cheese. \\
     \emph{Person1}: Do you have anything planned for today? \\
                ~~~~~~~~~~~~~~~ I think I am going to do some canning. \\
     \emph{Person2}: I am going to watch football. \\
                ~~~~~~~~~~~~~~~ What are you canning? \\
     \emph{Person1}: I think I will can some jam. \\
                ~~~~~~~~~~~~~~~ Do you also play footfall for fun? \\
     \emph{Person2}: If I have time outside of hunting \\
                ~~~~~~~~~~~~~~~ and remodeling homes. \\
                ~~~~~~~~~~~~~~~ Which is not much! \\
     \hline
  \noalign{\hrule height0.8pt} 
\end{tabular}
}
\end{center}
\caption{An example of Persona-Chat dataset}
\label{tab:personachat_case}
\end{table}

\begin{table*}[t!]
\begin{center}\small
\begin{tabular}{lll}
\hline
\noalign{\hrule height0.8pt} 
\textbf{Utterance}             & \textbf{Persona}            & \\ \hline
I do not. I spend all of my time playing frisbee.                   & I like playing frisbee.          & \\
Terrific. My favorite thing to do is dance.                         & I like dance.                    & \\
I love having hair on my face.                                      & I have a beard.                  & \\
You must like pizza as much as I like rush. You know? The band?     & I love the band rush.            & \\
I am a nurse. I work in the er and surgery unit.                    & I work at the hospital.          & \\
I do actually most performance art tho. Some abstract.              & I am an artist.                  & \\
I have a pet fish.                                                  & I have an exotic fish aquarium.  & \\
\hline
\noalign{\hrule height0.8pt} 
\end{tabular}
\end{center}
\caption{Examples of Persona-Link prediction}
\label{tab:persona-link_examples}
\end{table*}

\begin{table}[t!]
\begin{center}\small
\renewcommand\thetable{2}
{
\begin{tabular}{l}
\hline
  \noalign{\hrule height0.8pt}
    \textbf{Agent's personas:} \\
    My mom is a secretary. \\
    I am a bodybuilder. \\
    I have one brother. \\
    (-) I like to eat a lot of meat. \\
    (+) I am a meat eater. \\
    (+) My favorite meal is steak. \\
    \hline
    \noalign{\hrule height0.4pt}
    \textbf{Dialogue history:} \\
     \emph{User}: Hi there, how are you tonight? \\
     \emph{Agent}: Great. I just finished a huge steak. How are you? \\
     \emph{User}: I am good, drinking some scotch. \\
     \hline
    \textbf{Response without augmented personas:} \\
    \emph{Agent}: Cool! What are your hobbies?\\ 
    \hline
    \textbf{Response with augmented personas:}\\
    \emph{Agent}: I am major meat eater to build muscles.\\ 
  \hline
    \emph{User}: Oh, I like meat ok. \\
     \emph{Agent}: Scotch goes good with meat. \\
     \emph{User}: Scotch goes good with anything lol. \\
     \emph{Agent}: I have a brother who likes scotch. \\
     \emph{User}: I am an only child. \\
     \emph{Agent}: I have just the one brother and 3 sisters. \\
     \emph{User}: Do you have any pets? I am scared of dogs. \\
     \emph{Agent}: I have a cat and a dog.\\ You would not be scared of my dog. \\
     \emph{User}: What do you do for a living? \\
     \emph{Agent}: I am a secretary like my mom hard living and you? \\
     \emph{User}: I am an author. I write short stories. \\
     \emph{Agent}: Do you want to write a story \\~~~~~~~~~~~about a major meat eating bodybuilder? \\
     \hline
  \noalign{\hrule height0.8pt}
\end{tabular}
}
\end{center}
\caption{Full dialogue of Table~\ref{tab:chat_case}}
\label{tab:chat_case_full}
\end{table}

\subsection{\textsc{Persona-Link} model} As a Persona-Link model, we use \texttt{BERT-small} as a transformer for efficiency.
We use the Adam optimizer with a learning rate of 5e-5 with batch size 100, a max number of tokens of 64 for mention/persona encoder, epoch 10, weight decay 0.01, adam epsilon 1e-8, gradients clipping to 1, and distillation preference weight rate 1.0. The best parameters were chosen based on the Recall@10 score. Similar to \citet{mazare2018training}, during training we consider the other labels in the batch as negatives.

\subsection{\textsc{Persona-Chat} model}
We use the ParlAI implementation of Bi-encoder pre-trained on Reddit and fine-tuned on Persona-Chat as a dialogue agent. It is notable that the Bi-encoder for dialogue agent is different from that of Persona-Link model. While Persona-Link model links utterances to their referent personas, the dialogue agent predicts the next utterances in dialogue.

\subsection{Bi-encoder} As aforementioned, we use Bi-encoder as a model architecture of both Persona-Chat and Persona-Link. Here, we describe the architecture of Bi-encoder and its scoring methods. First, each context $c_{i}$ and corresponding $k$-th response candidate $r_{i,k}$ are encoded with transformers $T_{context}$ and $T_{response}$ respectively:
\begin{align}
    y_{c_{i}} = red(T_{context}(c_{i}))\nonumber\\
    y_{r_{i,k}} = red(T_{response}(r_{i,k}))\nonumber
\end{align}
where $red(\cdot)$ is a function that reduces the sequence of vectors into a vector. The output of [CLS] token is used as $red(\cdot)$ following~\cite{humeau2019poly}.
Then, the score of each response is computed by the dot-product:
\begin{equation*}
    s(c_{i}, r_{i,k}) = y_{c_{i}} \cdot y_{r_{i,k}}
\end{equation*}

In training time, the model is optimized by minimizing a cross-entropy loss. We use other correct responses in the same batch as negatives following~\cite{humeau2019poly} on training.

\section{Appendix B: Baselines of \textsc{Persona-Link} Task}\label{sec:appendix.3}

\begin{itemize}  \setlength\itemsep{0.1em}

    \item \textbf{Cosine Similarity:} This baseline selects a persona with the highest cosine similarity between the utterance and persona vectors using the pre-trained \texttt{word2vecf}~\cite{levy2014dependency} not word counts.

    \item \textbf{BM25}~\cite{robertson1995okapi}: A strong unsupervised retrieval model which is one of the most popular ranking functions in off-the-shelf search engines.
    
    \item \textbf{K-NRM}~\cite{Xiong_2017}: A kernel-based 
    efficient neural retrieval model that can consider both word-level soft/hard matching signals, \ie, semantic similarity via term embeddings and exact matches of query terms. 
    
    \item \textbf{Conv-KNRM}~\cite{dai2018convolutional}: A convolutional kernel-based neural ranking model that leverages n-gram soft matches.
 
    \item \textbf{Bi-encoder}~\cite{humeau2019poly}: 
    A state-of-the-art of transformer architecture for IR. We use BERT~\cite{devlin2018bert} as LM.
    \item \textbf{Paraphrasing}~\cite{mallinson2017paraphrasing}: Utterances and personas are first translated to a foreign pivot, then then back-translated to English, producing a paraphrase as a new sample.
    \item \textbf{EDA}~\cite{wei2019eda}: A powerful data augmentation tool for text classification with synonym replacement, random insertion, random swap, and random deletion.
\end{itemize}

\section{Appendix C: Examples}
\label{sec:appendix.2}

\subsection{An Example of \textsc{Persona-Chat} Dataset}
Table~\ref{tab:personachat_case} shows one of the dialogue in the Persona-Chat dataset. As shown, there are 4 persona sentences of \emph{Person2}, which \emph{Person2} speaker imitates. For example, in the first utterance of \emph{Person2}, \emph{Person2} says ``\emph{Hunting is one of my favorite hobbies.}'' imitating the persona ``\textbf{\texttt{I like to go hunting.}}''. 

\subsection{Examples of \textsc{Persona-Link} Prediction}
Table~\ref{tab:persona-link_examples} shows the predictions by Persona-Link model. Once trained, Persona-Link model links each utterance to its referent persona. For example, given ``\emph{I am a nurse. I work in the er and surgery unit.}'' as an utterance, Persona-Link model predicts ``\textbf{\texttt{I work at the hospital.}}'' as a persona, which can be seen as appropriate and can be used for learning better persona-grounded dialogue agents.

\subsection{Full Dialogue of Table 3}
Table~\ref{tab:chat_case_full} in this section shows a full dialogue of Table~\ref{tab:chat_case} with the top ranked response by the dialog agent. Original personas, removed persona (-), augmented personas (+) by a linking model, dialogue history, and two responses without and with augmented personas are given in each slot. 

That is, Table~\ref{tab:chat_case_full} presents retrieved response samples with comparison between without and with augmented personas. While a persona ``\textbf{\texttt{I like to eat a lot of meat.}}'' is removed, Persona-Link model can augment additional personas, such as ``\textbf{\texttt{I am a meat eater.}}'', which supplement missing information. As a result, although the inferred personas are a little bit different with the original one, the dialogue agent model with them correctly responds to the user's question, but the response without augmented personas is not the case.

\end{document}